\documentclass{article}

\usepackage[preprint]{neurips_2021}

\usepackage[utf8]{inputenc} 
\usepackage[T1]{fontenc}    
\usepackage{hyperref}       
\usepackage{url}            
\usepackage{booktabs}       
\usepackage{amsfonts}       
\usepackage{nicefrac}       
\usepackage{microtype}      
\usepackage{xcolor}         

\usepackage{natbib}
\usepackage{commath}

\usepackage[english]{babel}
\usepackage{amssymb,amsthm,amsmath}
\usepackage{enumitem}

\hypersetup{colorlinks=true,citecolor=}

\theoremstyle{plain}

\theoremstyle{definition}

\theoremstyle{remark}

\usepackage{newcommand}
\usepackage{graphicx, caption, subcaption, comment}

\title{Vision Transformer Compression with Structured Pruning and Low Rank Approximation}

\author{%
  Ankur Kumar   \\
  Department of Computer Science\\
  University of California, Los Angeles\\
  \texttt{ankurkr@ucla.edu}\\
}

\begin{document}

\maketitle

\begin{abstract}
Transformer architecture has gained popularity due to its ability to scale with large dataset. Consequently, there is a need to reduce the model size and latency, especially for on-device deployment. We focus on vision transformer proposed for image recognition task \citep{dosovitskiy2021image}, and explore the application of different compression techniques such as low rank approximation and  pruning for this purpose. Specifically, we investigate a structured pruning method proposed recently in \cite{Zhu2021VisionTP} and find that mostly feedforward blocks are pruned with this approach, that too, with severe degradation in accuracy. We propose a hybrid compression approach to mitigate this where we compress the attention blocks using low rank approximation and use the previously mentioned pruning with a lower rate for feedforward blocks in each transformer layer. Our technique results in 50\% compression with 14\% relative increase in classification error whereas we obtain 44\% compression with 20\% relative increase in error when only pruning is applied. We propose further enhancements to bridge the accuracy gap but leave it as a future work.
\end{abstract}

\section{Introduction}
BERT \citep{Devlin2019BERTPO} and GPT \citep{Radford2018ImprovingLU} inspired a trend of pre-training huge neural networks on large scale dataset for very accurate results on downstream tasks. Large pre-training corpus and a model that scales well on such dataset are critical in this process. \cite{Kaplan2020ScalingLF} and \cite{Henighan2020ScalingLF} show that transformer architecture \citep{Vaswani2017AttentionIA} strikes a balance between computation efficiency and scalability. Consequently, training transformer models with several billion parameters has become a trend to achieve high accuracy \citep{Ramesh2021ZeroShotTG, Radford2021LearningTV}. Vision transformer \citep{dosovitskiy2021image} show that large scale pre-training of transformers can benefit computer vision tasks also. The problem, however, is the resource constrained environment for deployment, specially for on device settings.

Model compression is a well researched topic. Several  techniques such as knowledge distillation, pruning, low rank approximation and quantization have been studied extensively to reduce model size without significantly degrading accuracy. However, few works have applied these compression techniques to vision transformer \citep{Zhu2021VisionTP, Yu2021AUP, Yang2021NViTVT, Chen2021ChasingSI, Hou2022MultiDimensionalMC}. Most of these works propose variations of structured pruning, which does not require specialize hardware to run the pruned model as opposed to unstructured pruning.

In this work, we follow \cite{Zhu2021VisionTP} because of their simplistic and structured pruning approach. However, we question their importance score estimation for pruning groups of neurons, and analyze the pruned network for this purpose. We conduct experiments on CIFAR-10 dataset to  provide empirical evidence that the pruning strategy may be indeed suboptimal. In this process, we come up with a hybrid compression technique where we apply low rank approximation in combination with the above pruning techniqe to mitigate some of the accuracy degradation. We further discuss that advanced techniques for pruning \citep{Molchanov2019ImportanceEF} and low rank approximation \citep{Chen2021DRONEDL} can be used in our \emph{modular} framework to further improve model performance.

Next, we present a brief overview of prior works on model compression, including compressing transformer models, in section \ref{related works}. We present our method in section \ref{methods}. Section \ref{experiments} and \ref{results} contain details of our experiments. Finally, we discuss shortcomings of the pruning technique mentioned earlier with possible future directions in section \ref{discussion}, and conclude in section \ref{conclusion}.

\section{Related Work} \label{related works}
\subsection{Knowledge Distillation}
\cite{Hinton2015DistillingTK} proposed to train a smaller neural network by first training a much bigger model and then training the smaller one using soft probabilities of the previous one. Since then, this technique has been successfully applied in multiple applications \citep{Gou2021KnowledgeDA}. However, we leave out this type of compression technique since it involves training the \emph{teacher} model first and then running it again to train a \emph{student} model, which may require significant computational resources.

\subsection{Quantization}
A simple way to compress a model is to use low precision for weights and activations. With 8-bit quantization, model size can be reduced by a factor of 4 with negligible accuracy loss \citep{Krishnamoorthi2018QuantizingDC, Nagel2021AWP}. Multiple efforts try to further reduce the precision, train a model with low precision etc \citep{Gupta2015DeepLW}. However, it is difficult to lower the precision beyond 16-bit without significant accuracy degradation and specialized hardware \citep{Wang2018TrainingDN}. It has become a trend to use half precision for mixed precision training to avoid the accuracy drop and make use of the available hardware \citep{Micikevicius2018MixedPT}.

\subsection{Pruning}
Over-parameterization of neural networks is considered to be one of the reasons behind its superior performance. Many works have also established that a lot of \emph{unimportant} weights can be removed to compress the models without sacrificing accuracy \citep{Han2015LearningBW, Blalock2020WhatIT}. There are several criteria to decide the importance score \citep{LeCun1989OptimalBD, Hassibi1992SecondOD, Molchanov2019ImportanceEF}. A common and convenient way is to zero out the weights with small magnitude. This is a type of \emph{unstructured} pruning. However, simply zeroing out does not improve the model size and speed unless there is specialized hardware. Another method, called \emph{structured} pruning \citep{Li2017PruningFF,Wen2016LearningSS}, avoids this issue by dropping a group of weights such as entire convolution filter or attention head. Therefore, we consider structured pruning in this work and combine it with other compression techniques \citep{Han2016DeepCC}.

\subsection{Low Rank Approximation}
Decomposition of weight matrices into low rank matrices is yet another technique to reduce the model size without affecting its structure. In case of convolution, factorizing channel and spatial convolution weights is used in place of traditional convolutional operation due to lesser number of parameters. Feedforward weight matrices can be factorized into two lower rank matrices to reduce the number of parameters \citep{Sainath2013LowrankMF}. However, it is not necessary that weight matrices are low rank, which we confirm for vision transformer from our experiments. The low rank structure can still be used if input feature is believed to lie in smaller dimensions \citep{Chen2021DRONEDL}.

\subsection{Transformer Compression}
With the success of BERT, several works have studied how to make transformer models smaller. DistilBERT \citep{Sanh2019DistilBERTAD} is a widely used compressed version of BERT models. Another line of work tries to make transformer models more efficient since computational complexity of transformer model grows quadratically in the number of input tokens. This is challenging for natural language processing, specially speech processing. \cite{Wang2020LinformerSW} show that attention matrix is typically low rank, and therefore, can benefit from low rank approximation.

Most of the works in computer visoin domain focus on removing uninformative patch tokens to speed up the training and inference steps \citep{Wang2021NotAI}. Vision transformer pruning (VTP) \citep{Zhu2021VisionTP} removes unimportant dimensions (columns or rows) of matrices in a transformer block. Inspired by several works combining multiple compression strategies \citep{Han2016DeepCC, Mao2020LadaBERTLA}, we build upon VTP and first show the shortcomings of this technique. Then we propose to use low rank approximation in combination with VTP to mitigate some of the issues. Our work is orthogonal to token pruning, quantization and knowledge distillation and can be combined with these approaches. 

\section{Method} \label{methods}
\subsection{Vision Transformer}
\begin{figure}[h]
    \centering
    \begin{subfigure}{0.45\textwidth}
        \includegraphics[scale=0.4]{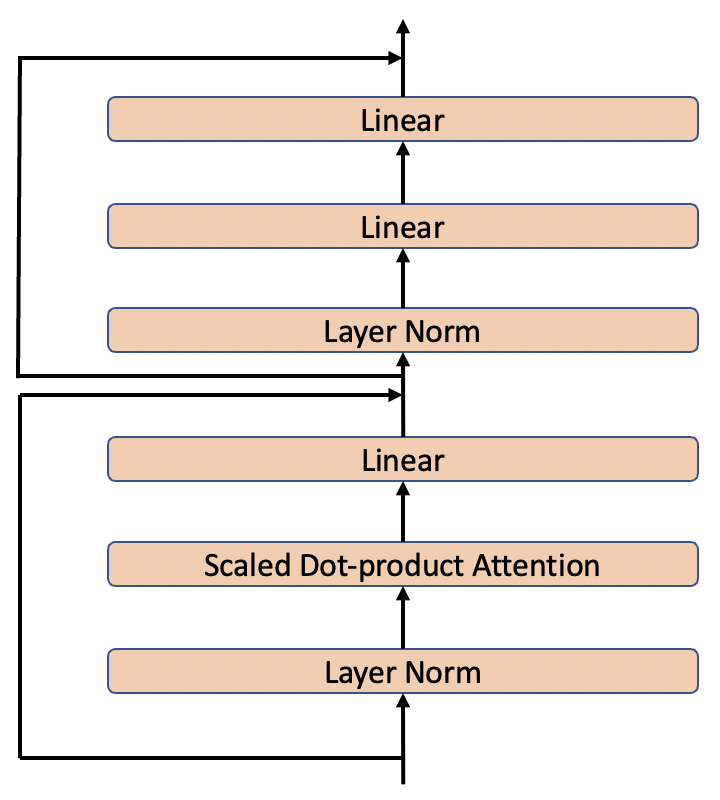}
        \subcaption{Transformer block}
        \label{transformer}
    \end{subfigure}
    \begin{subfigure}{0.4\textwidth}
        \includegraphics[scale=0.4]{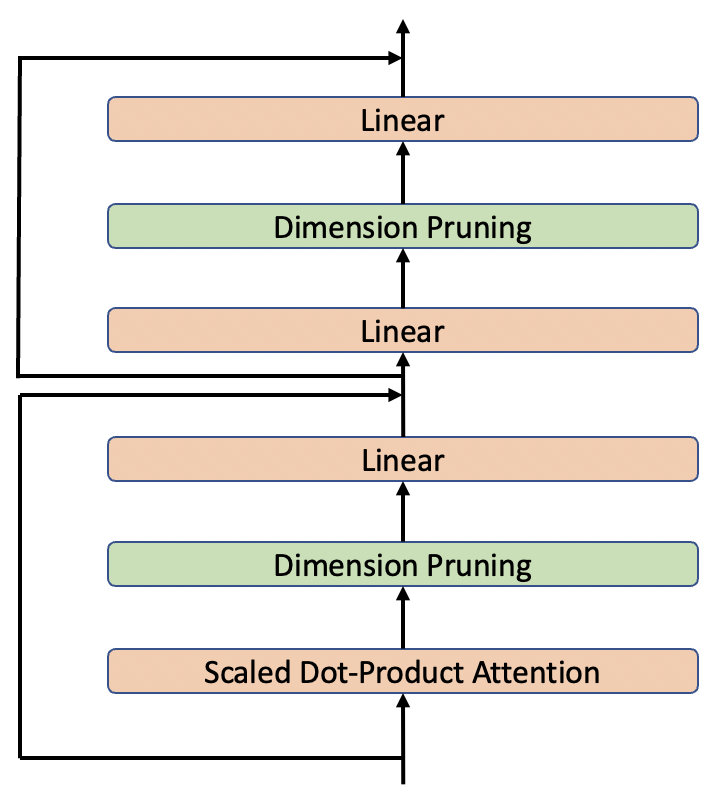}
        \subcaption{VTP (norm layer not shown)}
        \label{transformer_vtp}
    \end{subfigure}
    \caption{Transformer block with and without pruning layers. GELU non-linearlity is not shown in the feedforward block (consisting of top two linear layers) but applied on top of the linear layer outputs. We do not show the normalization layer in pruning case for simplicity.}
\end{figure}

An input image $x \in \mathbb{R}^{H \times W \times C}$ is divided and reshaped to have $P \times P$ patches, each having $P^2C$ pixels. The reshaped input $x_P$ has shape $\mathbb{R}^{N \times P^2C}$ where $N = HW/P^2$. Each patch is linearly projected into a $d$ dimensional space using a learnable embedding matrix. Following BERT, a learnable embedding corresponding to special token $[class]$ is prepended to $x_P$ making total sequence length $N+1$. This 1-d sequence of $d$ dimensional token can be fed to standard transformer model. \cite{dosovitskiy2021image} use learnable 1-d positional encoding which is added to the token embeddings. Structure of a transformer block is as shown in figure \ref{transformer}. We do not show GELU non-linearity which is applied after each of the top two linear layers in the block.

Output embedding corresponding to the $[class]$ token after the final transformer block serves as the input image representation. A classification head is attached on top of this to train the model. The head may contain additional hidden layer.

\subsection{Vision Transformer Pruning}
Each transformer block has almost all of the parameters in weight matrices of linear layer and query, key, value of self-attention layers. Vision transformer pruning \cite{Zhu2021VisionTP} aims to drop unimportant columns or rows from a matrix. To achieve this, dimension pruning layers are inserted between two layers containing weight matrices as shown in figure \ref{transformer_vtp}. The pruning layer is a binary mask with dimension equal to the input dimension of next layer (or output dimension of previous layer) and is multiplied with the output of previous layer. Therefore, if $i^{th}$ entry in the mask is 0, then it is equivalent to dropping $i^{th}$ column from the next layer and $i^{th}$ row from the previous layer. If the previous layer happens to be a self-attention layer, then $i^{th}$ row is dropped from each of the key, query and value matrices.

Since binary mask is non-differentiable, the above method relaxes mask to be a continuous value during training. More precisely, all mask values in a dimension pruning layer is initialized to 1 indicating that initially no column or row is dropped at the beginning. The following loss is optimized to learn the weights and mask
\begin{align*}
    L_{CE} + \lambda \sum_{l=1}^L |m_l|_{L1}
\end{align*}
where $L_{CE}$ is the regular cross entropy loss for classification task and $|m_l|_{L1}$ is the L1 norm of the mask or dimenison pruning layer in $l^{th}$ transformer block. Intuitively, the L1 loss promotes sparsity making some of the mask values close to 0, which results in lower contribution of corresponding dimensions in the overall activation. Therefore, we can drop the entire column to obtain less parameters. During inference, we again discretize the mask as follows
\begin{align*}
    m_l = m_l > \tau
\end{align*}
where $\tau$ is the threshold determined from desired pruning rate.

\subsection{Hybrid Compression Framework} \label{hybrid_method}
\begin{figure}[htp]
    \centering
    \subfloat[]{%
        \includegraphics[height=0.23\textheight]{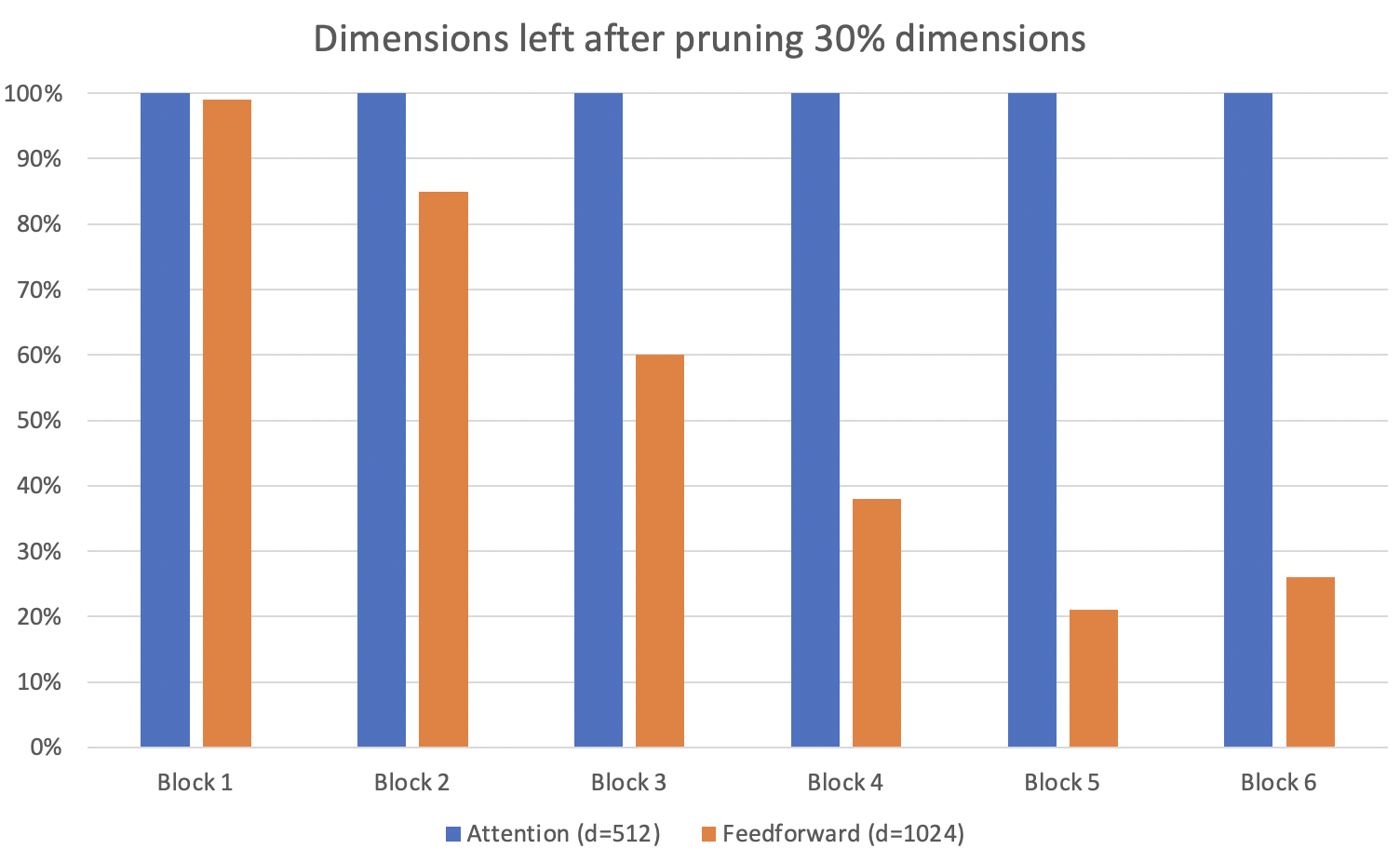}%
        \label{vtp_attn_ffn_prune}
    }%
    \subfloat[]{%
        \includegraphics[height=0.23\textheight]{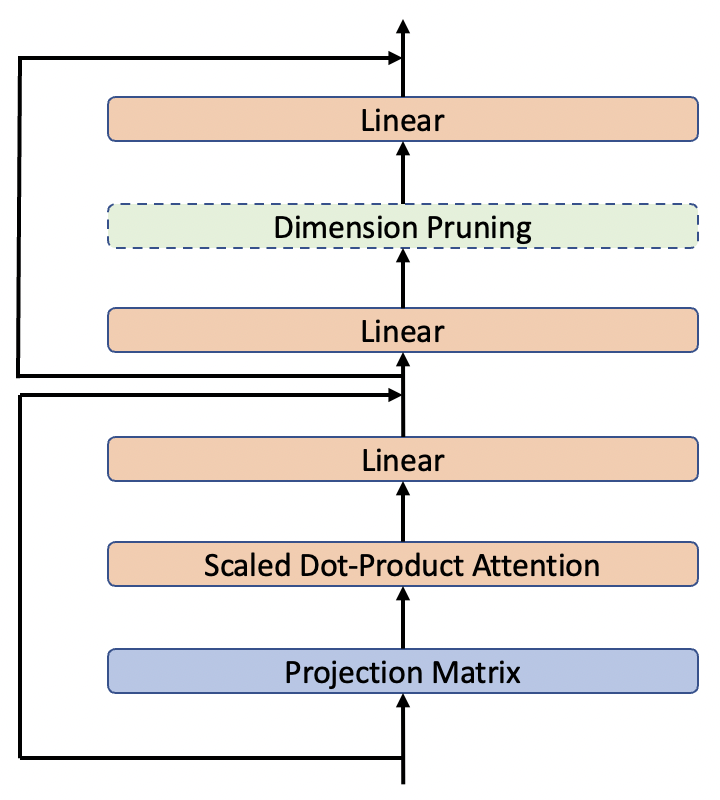}%
        \label{hybrid_pruning}
    }%
    \caption{(a) Effect of pruning with VTP in vision transformer layers. Feedforward layers are pruned and self-attention layers remain unaffected (b) Hybrid compression framework. Layer norm is not shown but applied before projection matrix and first linear layer in feedforward block}
    \label{hybrid_pruning}
\end{figure}

Figure \ref{vtp_attn_ffn_prune} shows the effect of applying VTP to vision transformer. We observe that mainly matrices from feedforward modules (top two linear layers in a transformer block) are pruned whereas attention weight matrices and projection layers are majorly unpruned. We discuss the reasons in more detail in section \ref{discussion}. However, this motivates us to compress self-attention block (query, key, value matrices in scaled dot-product layer and the following linear layer) in combination with above pruning approach for extra compression. We apply low rank approximation (LRA) for this purpose.

One issue with low rank approximation is that new rank $k$ must be much smaller than original rank $d$ since reducing the rank from $d$ to $k=d/2$ does not lead to lesser parameters if the original matrix has shape $d \times d$. We partly avoid this problem by applying LRA only to the attention block, where we share one of the low rank matrices among key, query and value matrices. It can be implemented by projecting $d-$dimensional input of a transformer block to a lower dimensional space of size $k$ by inserting a linear layer or projection matrix as shown by blue layer in figure \ref{hybrid_pruning}(b). This changes input dimension of query, key and value matrices from $d$ to $k$. We can now compute attention weights in either $d-$dimensional space or a lower $k-$dimensional space. We refer to the former as LRA($d \times k$) and the later as LRA($k \times k$). LRA($d \times k$) does not affect the size of linear layer after self-attention layer. However, LRA($d \times k$) reduces the input dimension of this layer also to $k$.

We apply LRA only to the attention block and not the feedforward block. This is because by design, transformer feedforward block is a factorized matrix where hidden dimension in between them is kept higher. We experiment with only LRA applied to vision transformer. We also apply VTP discussed earlier in the feedforward block (as shown in dashed green layer in figure \ref{hybrid_pruning}(b)) with LRA in attention block. This gives our hybrid compression framework.

\section{Experiments} \label{experiments}
We conduct experiments on CIFAR-10 dataset which has 10 output classes, 50,000 training images and 10,000 test images. We do not use any data augmentation. All the models are trained on a single gpu with a batch size of 64 and Adam optimizer with $10^{-4}$ learning rate. We also apply dropout of 0.1 to input feature embedding (after adding positional embedding) and all the weight matrices in the network except attention key, query and value matrices. Unless mentioned otherwise, we train for 15 epochs due to resource constraint. However, we also train a model for 50 epochs to validate our hypothesis which we discuss in section \ref{discussion}. Further, a pruned network is normally finetuned for few epochs to recover any accuracy drop due to pruning. However, we do not do finetuning because we believe that our model is not fully trained. As a result, it may be difficult to credit improvement in performance to either finetuning process or the fact that models are trained for extra epochs while comparison with the unpruned model.

\section{Results} \label{results}
\subsection{Vision Transformer Training}
We conduct hyper-parameter tuning for vision transformer for which results are shown in table \ref{vit}. We experiment with patch size $P$ of 2, 4 and 8. Results are shown in the first 3 rows. Patch size of 8 performs the worst and patch size of 2 performs the best. However, we choose patch size of 4 because of better training efficiency since transformer computational complexity grows quadratically with number of input tokens.

Then we varied the attention dimension and hidden dimension of feedforward blocks. Attention dimension of 512 and feedforward hidden dimension of 1024 worked the best as given in $6^{th}$ row. Finally, we changed the number of transformer blocks. We did not find significant accuracy difference between 6 and 9 layers. Therefore, we chose 6 transformer blocks in our network for better computation time. All the models compute self-attention with 8 heads (512/8 = 64 dimension for each head). Our final choice of hyper-parameter is shown in the second last row of table \ref{vit}. It achieves 72.61\% accuracy with 13M parameters on CIFAR-10 test set.

\begin{table}[t]
    \centering
    \begin{tabular}{cccccc}
        \toprule
        Patch Size & Dim & MLP Dim & Layers & Accuracy & Parameters \\
        \midrule
        2 (16x16) & 512 & 512 & 6 & 72.81\% & 10M \\
        4 (8x8) & 512 & 512 & 6 & 70.68\% & 10M \\
        8 (4x4) & 512 & 512 & 6 & 63.93\% & 10M \\
        \midrule
        4 (8x8) & 256 & 512 & 6 & 66.91\% & 3.5M \\
        4 (8x8) & 512 & 512 & 6 & 70.68\% & 10M \\
        4 (8x8) & 512 & 1024 & 6 & 72.61\% & 13M \\
        4 (8x8) & 512 & 1536 & 6 & 72.06\% & 16.6M\\
        \midrule
        4 (8x8) & 512 & 1024 & 3 & 68.64\% & 7M\\
        4 (8x8) & 512 & 1024 & 6 & 72.61\% & 13M\\
        4 (8x8) & 512 & 1024 & 9 & 73.03\% & 19.5M\\
        \bottomrule
    \end{tabular}
    \vspace{1mm}
    \caption{Hyper-parameter tuning for vision transformer trained on CIFAR-10 dataset}
    \label{vit}
\end{table}

\subsection{Vision Transformer Pruning} \label{vtp_results}
For pruning using VTP, we again train vision transformer with the final choice of hyperparameters. We insert the dimension pruning layers in each transformer block as shown in figure \ref{transformer}(b). We have only one hyper-parameter $\lambda$, which measures the strength of the sparsity or L1 loss. We experimented with values of $ 10^{-2}, 10^{-3} \text{ and } 10^{-4}$. We found that higher values of $\lambda$ significantly degraded the model performance but $10^{-4}$ worked well with a very small drop in model accuracy. This value is also the same as that in the original work although for ImageNet experiments. We show the results for VTP with different pruning rate in table \ref{vtp}. Prune rate of $\alpha$ means we want to zero out $\alpha$ fraction of all the mask values combined across all the pruning layers inserted in our model. Note that we have to re-train only one model with pruning layers as pruning rate is specified only at the inference time.
\begin{table}[t]
    \centering
    \begin{tabular}{ccccc}
        \toprule
        Model & Test Accuracy & Increase in Error & Model Size & Compression \\
        \midrule
        Base & 72.26\% & - & 13.2M (51MB) & - \\
        VTP (prune rate = 0.1) & 68.4\% & 14\%  & 12.3M (47MB) &  7\% \\
        VTP (prune rate = 0.3) & 69.4\% & 10\%  & 10.4M (40MB) &  21\% \\
        VTP (prune rate = 0.5) &  69.06\% & 11.5\% & 8.5M (33MB) & 36\%\\
        VTP (prune rate = 0.6) &  66.57\% & 20.5\% & 7.4M (29MB) & 44\% \\
        \bottomrule
    \end{tabular}
    \vspace{1mm}
    \caption{Vision Transformer Pruning}
    \label{vtp}
\end{table}
We can see from the second last row that we can compress the model by 36\% with around 12\% increase in error. However, the error goes up significantly if we compress the model further. We make two observations in this table. First, prune rate of 0.1 has higher error than prune rate of 0.3 which is counter-intuitive. We discuss this in detail in section \ref{discussion}. Second, pruning rate is not exactly the same as fractions of weights pruned because pruning rate is specified for fraction of dimensions to be pruned and not weights. We also observed that prune rate greater than 0.6 gave error perhaps because all dimensions of some weight matrix was getting dropped. This is indeed possible with VTP whereas low rank approximation in our method systematically compresses all attention blocks.

\subsection{Hybrid Compression}
We chose $k = d/4$ for low rank compression. The results for LRA are shown in table \ref{hybrid_results}. We note that we train a new model for all the LRA experiments (with and without pruning). We observe that only applying LRA is better than VTP for similar compression. For example, LRA $(d \times k)$ compresses the model by 24\% and has 5.8\% increase in error whereas pruning has 10\% increase in error for similar compression. Similarly, LRA $(k \times k)$ has 39\% compression with 8\% increase in error whereas pruning has 11.5\% increase in error. When we use pruning in feedforward block and LRA in attention block (hybrid compression), we achieve 50\% compression with 14\% increase in error whereas pruning by itself has 20.5\% increase in error for only 44\% compression, which is about 2\% absolute accuracy gap.

\begin{table}[h]
    \centering
    \begin{tabular}{ccccc}
        \toprule
        Model & Test Accuracy & Increase in Error & Model Size & Compression \\
        \midrule
        Base &  72.26\% & - & 13.2M (51MB) & - \\
        VTP & 69.06\% & 11.5\%  & 8.5M (33MB) &  36\% \\
        VTP &  66.57\% & 20.5\% & 7.4M (29MB) & 44\% \\
        \midrule
        LRA ($d$ x $k$) &  70.65\% & 5.8\% & 10M (39MB)  &  24\%\\
        LRA ($k$ x $k$) &  70.03\% & 8\% & 8M (31MB)  &  39\%\\
        VTP (FF only) + LRA ($d$ x $k$) & 68.46\% & 14\% & 6.6M (25MB) & 50\%\\
        \bottomrule
    \end{tabular}
    \vspace{1mm}
    \caption{Hybrid Compression for Vision Transformer}
    \label{hybrid_results}
\end{table}

\section{Discussion} \label{discussion}
\begin{figure}[t]
    \centering
    \subfloat[Mask values in attention blocks]{%
        \includegraphics[height=0.23\textheight]{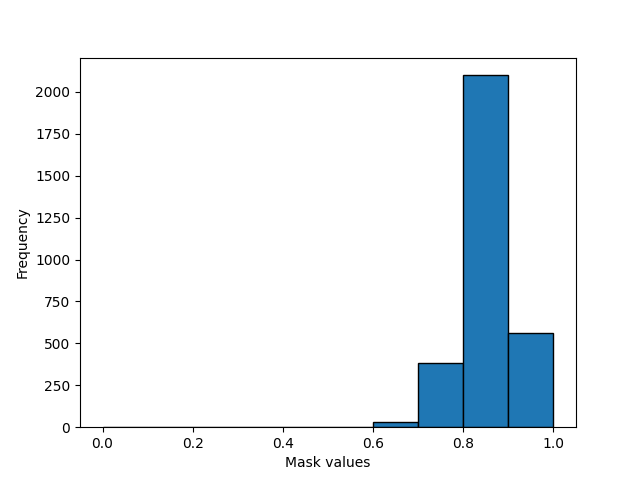}%
    }%
    \subfloat[Mask values in feedforward blocks]{%
        \includegraphics[height=0.23\textheight]{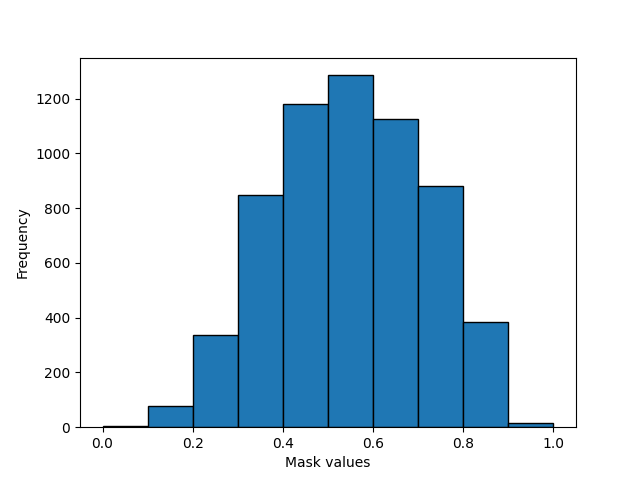}%
    }%
    \caption{Histogram of all mask values in attention and feedforward blocks after training vision transformer with pruning}
    \label{mask_values}
\end{figure}

In this section, we discuss multiple shortcomings of VTP approach. The first issue, mentioned in section \ref{hybrid_method}, is the type of layers getting pruned. Specifically, mostly dimensions for feedforward weight matrices are dropped and attention blocks are relativley unpruned as shown in figure \ref{hybrid_pruning}. This means that mask values for attention blocks would be much higher as compared to mask values for feedforward block since smaller mask values are pruned. We plot the histogram of mask values for attention blocks and feedforward blocks separately as show in figure \ref{mask_values}. It shows that, indeed, most attention mask values are greater than 0.8 whereas feedforward mask values roughly follow a normal distribution around 0.5.  

The second issue as mentioned in section \ref{vtp_results} is VTP having lower accuracy with prune rate of 0.1 as compared to prune rate of 0.3. One reason for this could be that mask values are not close to 0 after training for 15 epochs. However, we hypothesize that mask values after training may not actually reflect the importance of corresponding dimensions. So, while pruning, we may be removing an important feature dimension whose absolute mask value is the lowest, resulting in significant drop in accuracy (72.26\% to 68.4\%). In addition, if the dimensions being pruned were indeed of lowest importance, the accuracy should not have improved to 69.4\% on increasing the pruning rate to 0.3.

\begin{figure}[htp]
    \centering
    \subfloat[Mask values in attention blocks]{%
        \includegraphics[height=0.23\textheight]{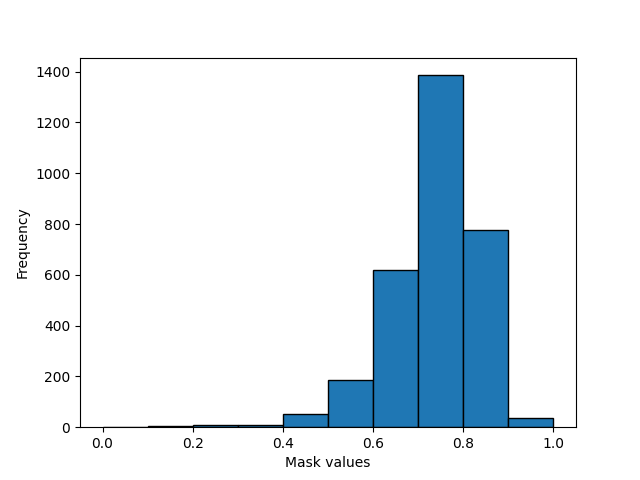}%
    }%
    \subfloat[Mask values in feedforward blocks]{%
        \includegraphics[height=0.23\textheight]{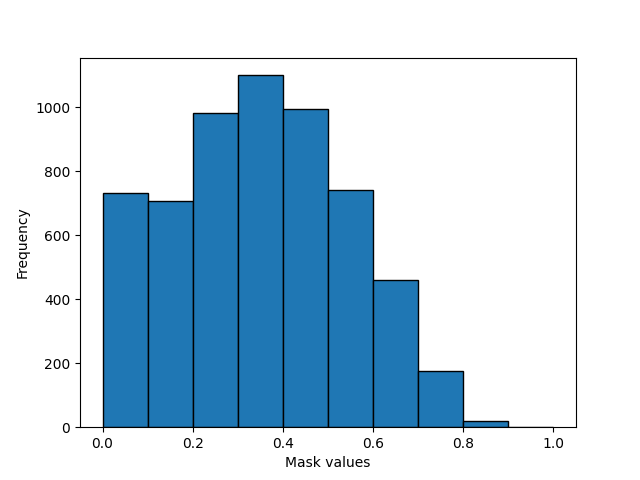}%
    }%
    \caption{Histogram of all mask values in attention and feedforward blocks after training vision transformer with pruning for \textbf{50 epochs}}
    \label{mask_values50}
\end{figure}

We verify our hypothesis that importance of a dimension may not be correlated with importance of corresponding mask values by training a model for 50 epochs. We expect many mask values to be close to 0 this time due to L1 loss and longer training. The distribution of mask values are shown in figure \ref{mask_values50}. 
We find that mask values in attention blocks are not affected much. But many mask values for feedforward blocks indeed become 0 or close to 0. Therefore, we would expect that VTP should perform well for this case. It turns out that the accuracy drop here is even more severe with at least 48\% increase in error as shown in table \ref{vtp50}. We still observe the irregular pattern that accuracy increases and then decreases upon increasing the pruning rate from 0.1 to 0.6. This is because of hard thresholding to get binary mask values during inference as compared to soft mask values during training. For a small prune rate such as 0.1 or less, we may be converting many mask values, which are close to 0, to 1 resulting in a severe degradation in accuracy.
\begin{table}[t]
    \centering
    \begin{tabular}{ccccc}
        \toprule
        Model & Test Accuracy & Increase in Error & Model Size & Compression \\
        \midrule
        Base & 80.71\% & - & 13.2M (51MB) & - \\
        VTP (prune rate = 0.1) & 56.11\% & 127.5\%  & 12.3M (47MB) &  7\% \\
        VTP (prune rate = 0.3) & 62.48\% & 94.5\%  & 10.4M (40MB) &  21\% \\
        VTP (prune rate = 0.5) &  71.45\% & 48\% & 8.5M (33MB) & 36\%\\
        VTP (prune rate = 0.6) &  70.26\% & 54.2\% & 7.4M (29MB) & 44\% \\
        \bottomrule
    \end{tabular}
    \vspace{1mm}
    \caption{Vision Transformer Pruning (\textbf{50 epochs})}
    \label{vtp50}
\end{table}

This brings to one of the future directions. We would like to augment or replace VTP with pruning tecniques that actually consider the importance of a group of weights (dimension in VTP is one type of grouping) for final accuracy. For example, Taylor pruning \citep{Molchanov2019ImportanceEF} estimates the importance using first order Taylor approximation of error when a group of weights is dropped. It also removes the need to threshold mask values during inference due to which we observed severe degradation in accuracy for VTP.

Second direction that we would like to explore is improving the low rank approximation. We observe more than 2\% absolute drop in accuracy when we use LRA (row 4 and 5 in table \ref{hybrid_results}). This is a significant drop for compression of 40\% and a model which is far from the best accuracy that can be achieved. \cite{Chen2021DRONEDL} propose a data-driven approach for low rank approximation, although in natural language processing, and show that the method is provably optimal. We can benefit from the modularity of our compression method and apply both the enhancements for better model performance. We get the extra benefit that Taylor pruning and data-driven LRA can be readily applied to any pre-trained models, thereby eliminating the need to re-train a model from scratch as in current scenario.

\section{Conclusion} \label{conclusion}
We presented a hybrid compression framework for vision transformer using structured pruning and low rank approximation. The proposed LRA has better compression and accuracy tradeoff as compared to VTP. The combined or hybrid approach had even better compression with higher accuracy than VTP. We also discussed a major shortcoming of VTP approach and validated it on CIFAR-10 dataset. Finally, we proposed various ways in which the hybrid compression technique can be further improved using advanced pruning and low rank approximation techniques.

\bibliography{references}

\begin{thebibliography}{}

\bibitem[Blalock et~al., 2020]{Blalock2020WhatIT}
Blalock, D.~W., Ortiz, J. J.~G., Frankle, J., and Guttag, J.~V. (2020).
\newblock What is the state of neural network pruning?
\newblock {\em ArXiv}, abs/2003.03033.

\bibitem[Chen, 2021]{Chen2021DRONEDL}
Chen, P.~H. (2021).
\newblock Drone: Data-aware low-rank compression for large nlp models.

\bibitem[Chen et~al., 2021]{Chen2021ChasingSI}
Chen, T., Cheng, Y., Gan, Z., Yuan, L., Zhang, L., and Wang, Z. (2021).
\newblock Chasing sparsity in vision transformers: An end-to-end exploration.
\newblock {\em ArXiv}, abs/2106.04533.

\bibitem[Devlin et~al., 2019]{Devlin2019BERTPO}
Devlin, J., Chang, M.-W., Lee, K., and Toutanova, K. (2019).
\newblock Bert: Pre-training of deep bidirectional transformers for language
  understanding.
\newblock {\em ArXiv}, abs/1810.04805.

\bibitem[Dosovitskiy et~al., 2021]{dosovitskiy2021image}
Dosovitskiy, A., Beyer, L., Kolesnikov, A., Weissenborn, D., Zhai, X.,
  Unterthiner, T., Dehghani, M., Minderer, M., Heigold, G., Gelly, S.,
  Uszkoreit, J., and Houlsby, N. (2021).
\newblock An image is worth 16x16 words: Transformers for image recognition at
  scale.

\bibitem[Gou et~al., 2021]{Gou2021KnowledgeDA}
Gou, J., Yu, B., Maybank, S.~J., and Tao, D. (2021).
\newblock Knowledge distillation: A survey.
\newblock {\em ArXiv}, abs/2006.05525.

\bibitem[Gupta et~al., 2015]{Gupta2015DeepLW}
Gupta, S., Agrawal, A., Gopalakrishnan, K., and Narayanan, P. (2015).
\newblock Deep learning with limited numerical precision.
\newblock In {\em ICML}.

\bibitem[Han et~al., 2016]{Han2016DeepCC}
Han, S., Mao, H., and Dally, W.~J. (2016).
\newblock Deep compression: Compressing deep neural network with pruning,
  trained quantization and huffman coding.
\newblock {\em arXiv: Computer Vision and Pattern Recognition}.

\bibitem[Han et~al., 2015]{Han2015LearningBW}
Han, S., Pool, J., Tran, J., and Dally, W.~J. (2015).
\newblock Learning both weights and connections for efficient neural network.
\newblock {\em ArXiv}, abs/1506.02626.

\bibitem[Hassibi and Stork, 1992]{Hassibi1992SecondOD}
Hassibi, B. and Stork, D.~G. (1992).
\newblock Second order derivatives for network pruning: Optimal brain surgeon.
\newblock In {\em NIPS}.

\bibitem[Henighan et~al., 2020]{Henighan2020ScalingLF}
Henighan, T.~J., Kaplan, J., Katz, M., Chen, M., Hesse, C., Jackson, J., Jun,
  H., Brown, T.~B., Dhariwal, P., Gray, S., Hallacy, C., Mann, B., Radford, A.,
  Ramesh, A., Ryder, N., Ziegler, D.~M., Schulman, J., Amodei, D., and
  McCandlish, S. (2020).
\newblock Scaling laws for autoregressive generative modeling.
\newblock {\em ArXiv}, abs/2010.14701.

\bibitem[Hinton et~al., 2015]{Hinton2015DistillingTK}
Hinton, G.~E., Vinyals, O., and Dean, J. (2015).
\newblock Distilling the knowledge in a neural network.
\newblock {\em ArXiv}, abs/1503.02531.

\bibitem[Hou and Kung, 2022]{Hou2022MultiDimensionalMC}
Hou, Z. and Kung, S.~Y. (2022).
\newblock Multi-dimensional model compression of vision transformer.
\newblock {\em ArXiv}, abs/2201.00043.

\bibitem[Kaplan et~al., 2020]{Kaplan2020ScalingLF}
Kaplan, J., McCandlish, S., Henighan, T.~J., Brown, T.~B., Chess, B., Child,
  R., Gray, S., Radford, A., Wu, J., and Amodei, D. (2020).
\newblock Scaling laws for neural language models.
\newblock {\em ArXiv}, abs/2001.08361.

\bibitem[Krishnamoorthi, 2018]{Krishnamoorthi2018QuantizingDC}
Krishnamoorthi, R. (2018).
\newblock Quantizing deep convolutional networks for efficient inference: A
  whitepaper.
\newblock {\em ArXiv}, abs/1806.08342.

\bibitem[LeCun et~al., 1989]{LeCun1989OptimalBD}
LeCun, Y., Denker, J.~S., and Solla, S.~A. (1989).
\newblock Optimal brain damage.
\newblock In {\em NIPS}.

\bibitem[Li et~al., 2017]{Li2017PruningFF}
Li, H., Kadav, A., Durdanovic, I., Samet, H., and Graf, H.~P. (2017).
\newblock Pruning filters for efficient convnets.
\newblock {\em ArXiv}, abs/1608.08710.

\bibitem[Mao et~al., 2020]{Mao2020LadaBERTLA}
Mao, Y., Wang, Y., Wu, C., Zhang, C., Wang, Y., Yang, Y., Zhang, Q., Tong, Y.,
  and Bai, J. (2020).
\newblock Ladabert: Lightweight adaptation of bert through hybrid model
  compression.
\newblock In {\em COLING}.

\bibitem[Micikevicius et~al., 2018]{Micikevicius2018MixedPT}
Micikevicius, P., Narang, S., Alben, J., Diamos, G.~F., Elsen, E., Garc{\'i}a,
  D., Ginsburg, B., Houston, M., Kuchaiev, O., Venkatesh, G., and Wu, H.
  (2018).
\newblock Mixed precision training.
\newblock {\em ArXiv}, abs/1710.03740.

\bibitem[Molchanov et~al., 2019]{Molchanov2019ImportanceEF}
Molchanov, P., Mallya, A., Tyree, S., Frosio, I., and Kautz, J. (2019).
\newblock Importance estimation for neural network pruning.
\newblock {\em 2019 IEEE/CVF Conference on Computer Vision and Pattern
  Recognition (CVPR)}, pages 11256--11264.

\bibitem[Nagel et~al., 2021]{Nagel2021AWP}
Nagel, M., Fournarakis, M., Amjad, R.~A., Bondarenko, Y., van Baalen, M., and
  Blankevoort, T. (2021).
\newblock A white paper on neural network quantization.
\newblock {\em ArXiv}, abs/2106.08295.

\bibitem[Radford et~al., 2021]{Radford2021LearningTV}
Radford, A., Kim, J.~W., Hallacy, C., Ramesh, A., Goh, G., Agarwal, S., Sastry,
  G., Askell, A., Mishkin, P., Clark, J., Krueger, G., and Sutskever, I.
  (2021).
\newblock Learning transferable visual models from natural language
  supervision.
\newblock In {\em ICML}.

\bibitem[Radford and Narasimhan, 2018]{Radford2018ImprovingLU}
Radford, A. and Narasimhan, K. (2018).
\newblock Improving language understanding by generative pre-training.

\bibitem[Ramesh et~al., 2021]{Ramesh2021ZeroShotTG}
Ramesh, A., Pavlov, M., Goh, G., Gray, S., Voss, C., Radford, A., Chen, M., and
  Sutskever, I. (2021).
\newblock Zero-shot text-to-image generation.
\newblock {\em ArXiv}, abs/2102.12092.

\bibitem[Sainath et~al., 2013]{Sainath2013LowrankMF}
Sainath, T.~N., Kingsbury, B., Sindhwani, V., Arisoy, E., and Ramabhadran, B.
  (2013).
\newblock Low-rank matrix factorization for deep neural network training with
  high-dimensional output targets.
\newblock {\em 2013 IEEE International Conference on Acoustics, Speech and
  Signal Processing}, pages 6655--6659.

\bibitem[Sanh et~al., 2019]{Sanh2019DistilBERTAD}
Sanh, V., Debut, L., Chaumond, J., and Wolf, T. (2019).
\newblock Distilbert, a distilled version of bert: smaller, faster, cheaper and
  lighter.
\newblock {\em ArXiv}, abs/1910.01108.

\bibitem[Vaswani et~al., 2017]{Vaswani2017AttentionIA}
Vaswani, A., Shazeer, N.~M., Parmar, N., Uszkoreit, J., Jones, L., Gomez,
  A.~N., Kaiser, L., and Polosukhin, I. (2017).
\newblock Attention is all you need.
\newblock {\em ArXiv}, abs/1706.03762.

\bibitem[Wang et~al., 2018]{Wang2018TrainingDN}
Wang, N., Choi, J., Brand, D., Chen, C.-Y., and Gopalakrishnan, K. (2018).
\newblock Training deep neural networks with 8-bit floating point numbers.
\newblock In {\em NeurIPS}.

\bibitem[Wang et~al., 2020]{Wang2020LinformerSW}
Wang, S., Li, B.~Z., Khabsa, M., Fang, H., and Ma, H. (2020).
\newblock Linformer: Self-attention with linear complexity.
\newblock {\em ArXiv}, abs/2006.04768.

\bibitem[Wang et~al., 2021]{Wang2021NotAI}
Wang, Y., Huang, R., Song, S., Huang, Z., and Huang, G. (2021).
\newblock Not all images are worth 16x16 words: Dynamic vision transformers
  with adaptive sequence length.
\newblock {\em ArXiv}, abs/2105.15075.

\bibitem[Wen et~al., 2016]{Wen2016LearningSS}
Wen, W., Wu, C., Wang, Y., Chen, Y., and Li, H.~H. (2016).
\newblock Learning structured sparsity in deep neural networks.
\newblock In {\em NIPS}.

\bibitem[Yang et~al., 2021]{Yang2021NViTVT}
Yang, H., Yin, H., Molchanov, P., Li, H., and Kautz, J. (2021).
\newblock Nvit: Vision transformer compression and parameter redistribution.
\newblock {\em ArXiv}, abs/2110.04869.

\bibitem[Yu and Wu, 2021]{Yu2021AUP}
Yu, H. and Wu, J. (2021).
\newblock A unified pruning framework for vision transformers.
\newblock {\em ArXiv}, abs/2111.15127.

\bibitem[Zhu et~al., 2021]{Zhu2021VisionTP}
Zhu, M., Tang, Y., and Han, K. (2021).
\newblock Vision transformer pruning.

\end{thebibliography}

\end{document}